\documentclass[conference]{IEEEtran}
\IEEEoverridecommandlockouts

\usepackage{cite}
\usepackage{amsmath,amssymb,amsfonts}
\usepackage{algorithmic}
\usepackage{graphicx}
\usepackage{textcomp}
\usepackage{xcolor}
\usepackage{bm}
\usepackage{makecell}
\usepackage{color}

\usepackage{soul}
\usepackage[bold]{hhtensor}
\usepackage[american]{babel}
\usepackage[breaklinks=true,colorlinks,bookmarks=false]{hyperref}

\def\BibTeX{{\rm B\kern-.05em{\sc i\kern-.025em b}\kern-.08em
    T\kern-.1667em\lower.7ex\hbox{E}\kern-.125emX}}
    
\begin{document}

\title{A Coarse-to-Fine Instance Segmentation Network with Learning Boundary Representation\\

\thanks{$^*$ Bin-Bin Gao and Xiu Li are corresponding authors. This work was supported by the National Natural Science Foundation of China (Grant No. 41876098), the National Key R$\&$D Program of China (Grant No. 2020AAA0108303), and Shenzhen Science and Technology Project (Grant No. JCYJ20200109143041798). This work was done when Feng Luo was an intern at Tencent Youtu Lab.}
}

\makeatletter
\newcommand{\linebreakand}{%
  \end{@IEEEauthorhalign}
  \hfill\mbox{}\par
  \mbox{}\hfill\begin{@IEEEauthorhalign}
}
\makeatother
\iffalse
\author{
\IEEEauthorblockN{1\textsuperscript{st} Feng Luo}
\IEEEauthorblockA{\textit{Shenzhen International}\\
\textit{Graduate School} \\
\textit{Tsinghua University}\\
Shenzhen, China \\
luof19@mails.tsinghua.edu.cn}
\and
\IEEEauthorblockN{2\textsuperscript{nd} Bin-Bin Gao\IEEEauthorrefmark{\dag}}
\IEEEauthorblockA{\textit{Youtu Lab} \\
\textit{}\\
\textit{Tencent}\\
Shenzhen, China \\
csgaobb@gmail.com}
% \linebreakand
\and
\IEEEauthorblockN{3\textsuperscript{rd} Jiangpeng Yan}
\IEEEauthorblockA{\textit{Department of Automation} \\
\textit{}\\
\textit{Tsinghua University}\\
Beijing, China \\
yanjp17@mails.tsinghua.edu.cn}\\
\and
\IEEEauthorblockN{4\textsuperscript{th} Xiu Li\IEEEauthorrefmark{\dag}}
\IEEEauthorblockA{\textit{Shenzhen International}\\
\textit{Graduate School} \\
\textit{Tsinghua University}\\
Shenzhen, China \\
li.xiu@sz.tsinghua.edu.cn}
}
\fi

\author{Feng Luo$^{1}$,~Bin-Bin Gao$^{2*}$,~Jiangpeng Yan$^3$ and Xiu Li$^{1*}$ \\
\normalsize $^1$Shenzhen International Graduate School, Tsinghua University, Shenzhen, China\\
\normalsize $^2$Youtu Lab, Tencent, Shenzhen, China  \\
\normalsize $^3$Department of Automation, Tsinghua University, Beijing, China\\
{\tt\footnotesize {luof19@mails.tsinghua.edu.cn, csgaobb@gmail.com,  yanjp17@mails.tsinghua.edu.cn, li.xiu@sz.tsinghua.edu.cn}}
}

\maketitle

\begin{abstract}
Boundary-based instance segmentation has drawn much attention since of its attractive efficiency. However, existing methods suffer from the difficulty in long-distance regression. In this paper, we propose a coarse-to-fine module to address the problem. Approximate boundary points are generated at the coarse stage and then features of these points are sampled and fed to a refined regressor for fine prediction. It is end-to-end trainable since differential sampling operation is well supported in the module. Furthermore, we design a holistic boundary-aware branch and introduce instance-agnostic supervision to assist regression. Equipped with ResNet-101, our approach achieves 31.7\% mask AP on COCO dataset with single-scale training and testing, outperforming the baseline 1.3\% mask AP with less than 1\% additional parameters and GFLOPs. Experiments also show that our proposed method achieves competitive performance compared to existing boundary-based methods with a light-weight design and a simple pipeline. 
\end{abstract}


\section{Introduction}

Instance segmentation is a fundamental but challenging task in computer vision because it requires to predict all instances' categories, locations and shapes in an image simultaneously. This task is crucial for real-world scenarios such as robot grasping, autonomous driving, and video surveillance. In recent years, many approaches have been proposed to dealing with instance segmentation.

\begin{figure}[htbp]
\centerline{\includegraphics[scale=0.3, trim=0 0 0 0]{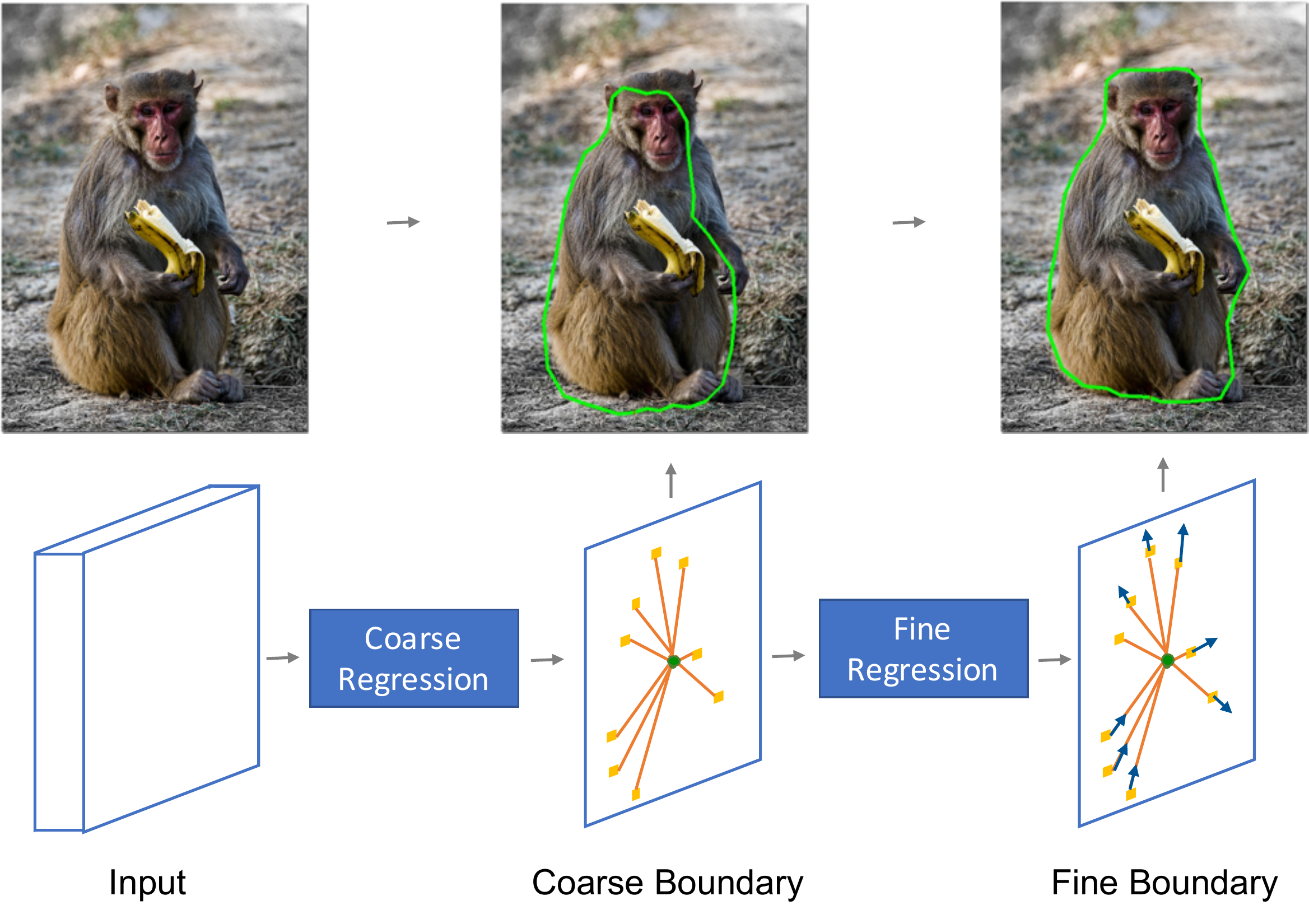}}
\caption{Boundary regression in a coarse-to-fine manner. The difficult long-distance regression is decomposed into coarse prediction (with a coarse regression module) in the global image and fine prediction (with a fine regression module) in the local area.}
\label{head}
\end{figure}

Existing works can be roughly divided into two families: pixel-based and boundary-based methods. Pixel-based methods~\cite{he2017mask,liu2018path,chen2019hybrid,huang2019mask} usually formulate the task as dense pixel-wise classification within each bounding box predicted by an object detector. Although achieving good performance, they are sensitive to inaccurate bounding boxes and suffer from heavy computation burden. Different from pixel-based instance segmentation, boundary-based methods \cite{liang2020polytransform,peng2020deep,xie2020polarmask,xu2019explicit} promote the dense pixel-wise classification to sparse point regression. Specifically, they usually represent an instance by a boundary point set and consider the task as boundary point regression. These methods do not need costly upsampling, thus leading to better efficiency. Besides, they are not limited to predicted bounding boxes and maybe achieve more robust performance.

Boundary-based instance segmentation predicts contour coordinates in different manners. Some works~\cite{peng2020deep,liang2020polytransform,cheng2019darnet,gur2020end} regress coordinates by gradually adjusting a set of initial points to targets. Their initial points are generated by a pre-trained detection network thus have a complex training pipeline. Other approaches~\cite{xie2020polarmask,xu2019explicit} model the boundary by one center and rays emitted from the center to the contour in polar coordinate. Especially, PolarMask~\cite{xie2020polarmask} designs an elegant single-shot instance segmentation framework and formulates the task as center classification and radius regression, which unifies object detection and segmentation. Viewing every pixel on the input image as a training example, it does not need anchors that are used in most works following the paradigm of ``detect then segment''. However, it is difficult to predict the fine instance boundary because of long-distance regression issue. In fact, a single feature vector of a pixel may not encode all information of boundary points that are far from the center because the response of input signal decays with distance from the center of the receptive field increasing~\cite{luo2016understanding}.

In order to solve the above issue, let us come back to how the human visual system works while observing objects. Given an image, we first glance through the whole image to locate objects' approximate positions. Guided by the coarse positions, we gradually focus on local areas around them and get a clear view of objects’ details. Beyond intuition, this coarse-to-fine visual processing mechanism is also supported by many neuroscience works~\cite{hegde2008time,peyrin2010neural,goffaux2011coarse,kauffmann2014neural}.

Inspired by the mechanism, we introduce a coarse-to-fine regression module and a holistic boundary-aware branch to tackle the problems in PolarMask~\cite{xie2020polarmask}. As illustrated in Fig.~\ref{head}, the coarse-to-fine regression module regresses radius via two steps. First, the coarse regression module generates approximate radius of sampled boundary points, which form an inaccurate boundary. Subsequently, the fine regression module corrects initial boundary points' radius with features sampled from specific locations where correspond to the coarse prediction to derive a precise contour. Furthermore, we design a holistic boundary-aware branch to take advantage of global image-level supervision. It predicts all boundary pixels without distinguishing instances, which is more coarse supervision. This branch is jointly trained with the regression branch and can be dropped at the inference stage, which brings no computation burden.

Experiments show that our approach equipped with ResNet-101~\cite{he2016deep} surpasses the PolarMask baseline ~\cite{xie2020polarmask} 1.3\% mask AP on COCO~\texttt{test-dev} with increasing less than 1\% parameters and less than 1\% GFLOPs. In addition, comparing with other boundary-based instance segmentation methods, our approach achieves competitive performance with single-scale training and testing.

In summary, our contributions are as follows:

\begin{itemize}
\item We propose a coarse-to-fine regression module to refine the radius of boundary points, which decomposes long-distance regression into two simple subtasks and thus effectively alleviates the difficulty of optimization in boundary-based instance segmentation.
\item We design a holistic boundary-aware branch and additionally introduce coarse instance-agnostic boundary supervision. It guides the feature extractor to learn global features and assists radius regression explicitly. 
\item Experimental results demonstrate that our two well-designed modules are effective and outperform the baseline 1.3\% mask AP with negligible parameters and computation overheads. 
\end{itemize}

\section{Related Work}
\subsection{Pixel-based Instance Segmentation} 
Most works~\cite{he2017mask,liu2018path,chen2019hybrid,huang2019mask} formulate instance segmentation as pixel-wise foreground-background classification within a region proposal. Mask R-CNN~\cite{he2017mask} adds an extra parallel branch on Faster-RCNN~\cite{ren2015faster} to segment instances within the predicted boxes. PANet~\cite{liu2018path} designs a bottom-up path augmentation to improve information flow, yielding better performance than Mask R-CNN.  Mask Scoring R-CNN~\cite{huang2019mask} proposes a mIOU head to regress the predicted mask's quality score, which avoids the inconsistency of the classification confidence map and the predicted mask quality. However, these methods require upsampling as post-procedure, which affects masks' quality negatively. Moreover, following the paradigm of ``detect then segment", they are time-consuming. Free of region proposals, other bottom-up methods~\cite{newell2017associative,de2017semantic,gao2019ssap,liu2017sgn} perform pixel-level semantic segmentation of the whole image and then cluster pixels into instances by auxiliary information, such as per-pixel embedding~\cite{de2017semantic} and affinity pyramid~\cite{gao2019ssap}. The accuracy of bottom-up methods~\cite{newell2017associative,de2017semantic,gao2019ssap,liu2017sgn} is generally inferior to top-down methods~\cite{he2017mask,liu2018path,chen2019hybrid,huang2019mask}.

\subsection{Boundary-based Instance Segmentation}
Boundary-based methods predict enclosed contours for instance segmentation. Some of them regress contours in a direct way. ExtremeNet\cite{zhou2019bottom} builds octagonal masks based on estimated extreme points of instances, which is a coarse outline.
ESE-Seg~\cite{xu2019explicit} explicitly encodes instances' shapes to a short vector via Chebyshev polynomial fitting. However, it is limited to the reconstruction error of the shape vector, especially at strict IOU thresholds. PolarMask~\cite{xie2020polarmask} exploits a mask head to directly regress radius of 36 boundary points with the same angle interval, which accomplishes single-shot instance segmentation. Unfortunately, its performance is still not perfect due to the difficulty of long-distance regression. In this paper, we introduce a lightweight coarse-to-fine module and a parallel holistic boundary-aware branch to make it more effective.

Influenced by active contour models, some works~\cite{liang2020polytransform,peng2020deep} predict boundary points by gradually deforming from initial boundaries. Nevertheless, they both have sophisticated architectures and cannot be trained end-to-end. PolyTransform~\cite{liang2020polytransform} exploits an additional strong segmentation network to generate initial segmentation masks and then polygons converted from the masks are fed to another deforming network to predict offsets of polygons' vertices. Similarly, Deep Snake~\cite{peng2020deep} employs CenterNet~\cite{zhou2019objects} to locate instances and then gradually deforms four midpoints of bounding boxes' edges to the target positions. During the deformation process, its deep snake modules need to iterate several times. To some extent, our method shares the ``gradually deforming" idea of \cite{liang2020polytransform,peng2020deep}, but our model is simpler and end-to-end trainable.

\begin{figure*}[htbp]
\centerline{\includegraphics[scale=0.4, trim=0 0 0 0]{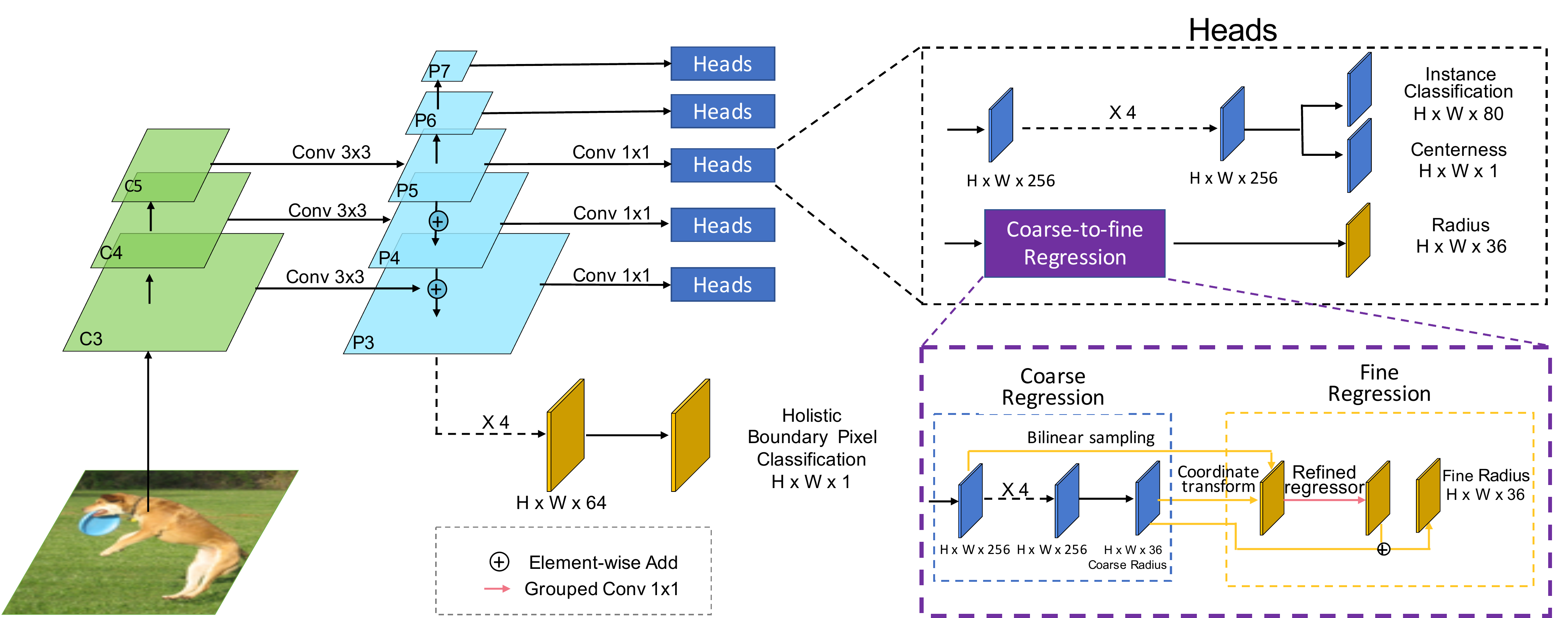}}
\caption{The overall structure of the proposed instance segmentation network. Building upon PolarMask~\cite{xie2020polarmask}, our framework adds a coarse-to-fine radius regression module and a holistic boundary-aware branch. The coarse regression uses PolarMask’s original radius head to generate the radius of boundary points. The fine module is composed of a coordinate transformation layer, a feature sampler and a refined regressor. The coordinate transformation layer transforms the coarse radius to Cartesian coordinates and then the feature sampler samples features of these coordinates from FPN feature maps using bilinear sampling. Finally, the refined regressor predicts for radius correction via a grouped Conv 1$\times$1, followed by an element-wise addition of the radius correction and the coarse radius. The holistic boundary-aware branch performs pixel-wise classification to predict contours in an instance-agnostic manner, which guides the network to extract global features. Here, $H$ and $W$ denote the height, width of feature maps, respectively.}
\label{architecture}
\end{figure*}

\section{Methods}
We aim at designing an elegant but efficient boundary-based instance segmentation network. To this end, we propose a coarse-to-fine module based on PolarMask~\cite{xie2020polarmask} to improve instance segmentation performance. Inspired by the human visual system, we firstly use PolarMask's original head to predict approximate boundary points through the whole image and then design a fine module to regress more accurate locations. Further, we develop a holistic boundary-aware branch to assist the boundary regression via global instance-agnostic supervision. This mechanism is similar to key-points prediction~\cite{tian2019directpose}.

In this section, we first introduce the boundary representation. Then we describe the proposed overall framework. Next, we clarify how the coarse-to-fine module works. Moreover, the holistic boundary-aware branch for incorporating global information is described. Finally, the loss function is briefly introduced.

\subsection{Boundary Representation}
An instance shape is conveniently represented by its boundary and the boundary is discretized to a set of points. We can represent these points by adopting Cartesian coordinates or Polar coordinates. However, Cartesian coordinates easily cause self-intersection since this parameterization follows separate functions for $x$ and $y$ coordinates~\cite{cheng2019darnet}. In order to avoid the self-intersection, we construct a polar coordinate to represent instances boundary following~\cite{xu2019explicit,cheng2019darnet,gur2020end,xie2020polarmask}.

Given an instance mask, the corresponding boundary contains two steps. First, we locate the mass center $p_c=(x_c, y_c)$ of this instance as the pole. Second, starting from the center $p_c$, $n$ rays are emitted uniformly with an equal polar angle interval $\Delta \theta$ from the pole to boundary points. In practice, 36 points are sampled to delineate an instance and $\Delta\theta$ is equal to $\frac{\pi}{18}$. To avoid self-intersection, we only sample the point with the largest radius if a ray insect with a boundary several times. 

With the fixed angle interval, boundary points possess an inherent order and can be described as a radius vector. Let ${(r_k, {\theta}_k)}_{k=1\cdots36}$ denotes polar coordinates of 36 sampled points, where $\theta_k = k\Delta\theta$. Then an instance can be represented as a mass center $p_c$ and the sample points’ radius vector $[r_1, \cdots, r_{36}]$. Our instance segmentation model can be formulated as, 
\begin{equation}
\Phi(I;w) = \{p_c, r_1, \cdots, r_{36}\},
\end{equation}
where $I$ is an input image and $\Phi(.;w)$ is a learned and parameterized convolutional neural network.

\subsection{Instance Segmentation Framework}

Following the network designed in~\cite{xie2020polarmask}, we propose a coarse-to-fine regression module and a boundary-aware branch to improve the boundary prediction quality of an instance. Meanwhile, it enjoys the simplicity of single-shot detection network~(FCOS)~\cite{tian2019fcos}. The overall structure of our instance segmentation network is shown in Fig.~\ref{architecture}.

The proposed framework consists of a backbone with a feature pyramid network~(\emph{FPN})~\cite{lin2017feature}, a holistic boundary-aware branch~(\emph{HBB}), a coarse-to-fine module~(\emph{CFM}) embedded on radius prediction head and another classification head presented in PolarMask. It takes a 3-channel image $I$ as input and applies a backbone network combined with FPN for extracting multi-scale features. Denote these five level feature maps as ${P_3,P_4,P_5,P_6,P_7}$. ${P_3, P_4,P_5}$ are produced by merging the bottom-up backbone feature maps (undergoing 1$\times$1 convolution to obtain the same channel number as that of the merging feature maps) and the corresponding upsampling feature maps in the top-down pathway. After merging, a 3$\times$3 convolution is applied to alleviate the upsampling aliasing. ${P_6,P_7}$ are both generated by convolving ${P_5,P_6}$ via 1$\times$1 kernel with stride 2.

Subsequently, multi-scale feature maps undergo two parallel heads for object classification and polar radius regression respectively, resulting in prediction with the same spatial size as input features. Specifically, these heads perform in a per-pixel fashion and every location on ${P_3,P_4,P_5,P_6,P_7}$ is viewed as a training sample. For each sample, the classification head with an additional parallel single-layer branch predicts an 80-dim vector of probabilities over 80 categories and a 1-dim centerness evaluating the distance from the current location to the center of its target object. At the test stage, a sample's confidence score is calculated by multiplying classification scores and predicted centerness to suppress the low-quality masks. The output of the radius regression head is the radius vector $[r_1, \cdots, r_{36}]$ with its pole at the current location. For each head, the parameters of the prediction head are shared among all feature levels, except a trainable scale scalar. For the largest feature map ${P_3}$, we additionally take it as input of holistic boundary-aware branch, which assists radius regression in the whole image context.

\subsection{Coarse-to-fine Regression module}

As discussed before, it is difficult to regress all boundary points based on the corresponding single feature vector in feature maps. To tackle the problem, we propose a coarse-to-fine radius regression head, which divides the process into coarse regression and fine prediction stage. Correspondingly, the head is composed of a coarse and a fine regression module. With the coarse-to-fine strategy, the feature vector for coarse prediction only needs to encode information of the targets' neighborhood roughly and the fine module would regress the precise locations in local areas after the initial prediction. The proposed coarse-to-fine regression module thus decomposes the difficult task into two simple subtasks.

In this paper, we use the PolarMask's original radius head as the coarse stage of our coarse-to-fine module. As shown in Fig.~\ref{architecture}, the fine module is attached to the coarse module and has three components: a coordinate transformation layer, a feature sampler and a refined regressor. Similarly, fine modules of five feature levels share the same weights but are with different trainable scale scalars to make the head parameter-efficient.

To clarify, a specific position $(x_c, y_c)$ of the coarse radius prediction on level $l$ is taken to explain how our fine module works. Let $r_{coarse}^k$ be the $k$-th coarse predicted radius of the position $(x_c, y_c)$ and $s$ be the stride of feature level $l$. First, a transformation layer is used to transform Polar coordinates to Cartesian ones. The process can be formulated as,
\begin{equation}
x_k=x_c+r_{coarse}^k\times\sin\theta_k /s,
\end{equation}
\begin{equation}
y_k=y_c+r_{coarse}^k\times\cos\theta_k /s,
\end{equation}
where $k=1, \cdots, 36$. The numerical range of the radius $r_{coarse}^k$ is adapted to the input image while sampling positions’ coordinates $(x_c, y_c)$ are associated with feature maps, so coordinates need rescaling by a factor of the feature level's stride $s$. Second, the feature sampler picks 36 feature vectors corresponding to $(x_k, y_k)$ from the feature map $P_l$ and concatenates them as a single vector $\vec f$. As sampling positions’ coordinates can be fractional, the bilinear sampling kernel in \cite{jaderberg2015spatial} is used. Finally, the refined regressor takes the feature vector $\vec f$ as input and predicts the refined radius of 36 sampled boundary points. In detail, the feature $\vec f$ is divided into 36 groups and a single grouped $1\times 1$ convolution is applied to regress each point’s refined radius separately, which achieves a similar accuracy with a standard convolution. Denote the refined regressor as $\Phi_{r}$, the final radius vector prediction can be formulated as 
\begin{equation}
\vec r = \vec r_{coarse}+ \Phi_r(\vec f).
\end{equation}

Note that the coarse prediction is not only used for feature extraction but is composed of our final predicted radius, which sets our work apart from Deformable Convolutional Networks (DCN)~\cite{dai2017deformable}. In addition, explicit supervision is applied to the coarse regression stage in our network, but it is infeasible in DCN because of the unavailable ground truth of offsets.

\subsection{Holistic Boundary-aware Branch}
Viewing each pixel as a training example, the coarse-to-fine regression module focuses on individual instances. Since shape complexity can be revealed better in the global scope of images \cite{wang2020centermask,kim2021devil}, we design a holistic boundary-aware branch and introduce image-level supervision. The branch performs instance-agnostic boundary pixel classification, which guides the network to encode holistic boundary information and assists radius regression explicitly. Taking a FPN feature map as input and applying four standard $3\times3$ convolutions with channels of 128, 64, 64, 64 successively, the branch outputs a boundary probability map with one channel. At the inference stage, the branch can be freely removed, which has no effect on computational complexity.

As shown in Fig.~\ref{reg_gt}, the branch's ground truth is a mask with boundary pixels being 1 and others 0. In detail, the boundary mask is generated as follows. Given masks of an image, we exploit the border following algorithm\cite{suzuki1985topological} to extract boundaries and draw all of them on a single mask. To match the size of the FPN feature map, for each boundary point $v\in V$, we compute a low-resolution equivalent $\widetilde{v}=\lfloor\frac{v}{s} \rfloor$.
Let $V$ denote the boundary point set and $m$ be the number of instances in the image. Formally,
\begin{equation}
V=\cup_{i=1}^{m}V_i=\cup_{i=1}^{m}\lbrace \widetilde{v}_{i1},...,\widetilde{v}_{iq^i} \rbrace,
\end{equation}
where $q^i$ denotes the boundary point number of the $i$-th instance. 
Empirically, low-resolution masks negatively affect regression performance with a bold boundary, while high-resolution boundary masks preserve more details and promote the regression module better. Therefore, the boundary-aware branch is only applied on the largest FPN feature map $P_3$ and $s$ is the stride of $P_3$. 

\begin{figure}[htbp]
\centering
\centerline{\includegraphics[scale=0.56,trim=0 0 0 0]{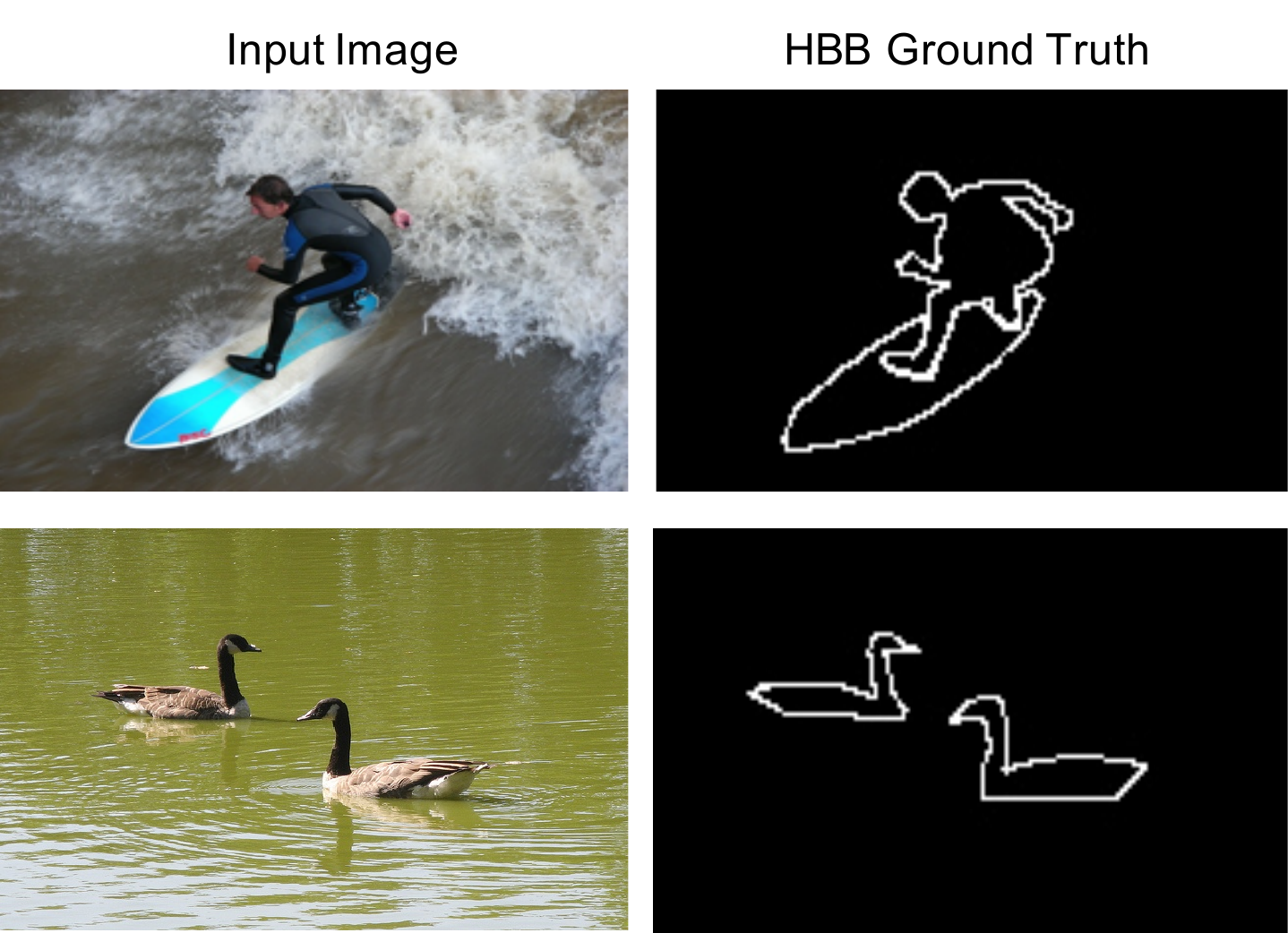}}
\caption{Examples of ground truth of our holistic boundary-aware branch. The boundary is instance-agnostic and its values are labeled as 1 and others 0.}
\label{reg_gt}
\end{figure}

\subsection{Loss Function}
We define our training loss function as follows:
\begin{equation}
    L=L_{cls}+L_{cnt}+\alpha L_{coarse}+L_{fine}+L_{hbb},
\label{loss}
\end{equation}
where $\alpha$ is a hyper-parameter and its effect is
discussed in ablation study~(Section~\ref{as}). Note that there are serious class imbalance issue when we perform holistic boundary pixel classification and object category classification. To alleviate the conflict, we employ focal loss~\cite{lin2017focal} for object classification~($L_{cls}$) and holistic boundary pixel classification~($L_{hbb}$). For coarse and fine polar coordinate regression, polar IOU loss~\cite{xie2020polarmask} is used in $L_{coarse}$ and $L_{fine}$. Like~\cite{tian2019fcos, xie2020polarmask}, the centerness~($L_{cnt}$) is trained by binary cross entropy loss. The learning of our framework is performed using stochastic gradient descent and standard back propagation.

\section{Experiments}
\subsection{Datasets and Evaluation Metrics}
We perform experiments on the most challenging instance segmentation dataset COCO\cite{lin2014microsoft}. Following the common practice\cite{ren2015faster,lin2017feature}, we use the union of 80k training images and a 35k subset of val images (\texttt{trainval35k}) for training. The remaining 5k val images (\texttt{minival}) are as the validation set for ablation study. COCO \texttt{test-dev} set is used for comparison with other methods.

The performance of predicted masks are evaluated by the average precision (AP, AP$_{50}$, AP$_{75}$, AP$_{S}$, AP$_{M}$, AP$_{L}$). AP is computed at ten IOU overlap thresholds ranging from 0.5 to 0.95 with a step size of 0.05. AP$_{50}$, AP$_{75}$ are AP at IOU 50\% and IOU 75\% respectively. AP$_{S}$, AP$_{M}$, AP$_{L}$ are AP for objects at different sizes\cite{lin2014microsoft}.

\begin{figure*}[htbp]
\centerline{\includegraphics[scale=0.5]{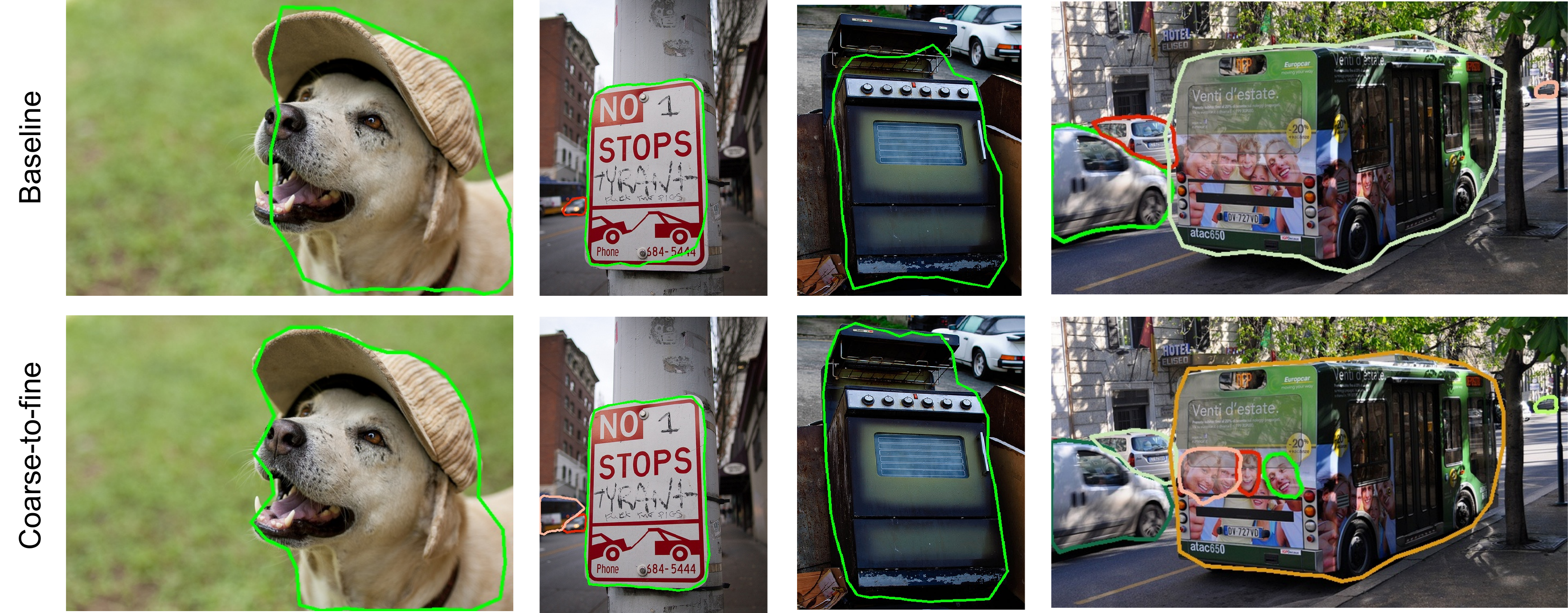}}
\caption{Comparison between coarse-to-fine regression and direct regression. Boundaries with different colors are used to differentiate multiple instances in the same image. As shown, the boundaries generated by the coarse-to-fine strategy are overall of higher quality. Besides, some instances missing in the baseline result can be captured by the coarse-to-fine module.}
\label{corase_to_fine_results}
\end{figure*}

\subsection{Implementation Details} 
Unless specified, ResNet-101~\cite{he2016deep} is used as the backbone of our network and the same hyper-parameters with PolarMask~\cite{xie2020polarmask} are used. Additionally, our method introduces nother two hyper-parameters, the focusing parameter $\gamma$ in $L_{hbb}$ and the coefficient $\alpha$ of $L_{coarse}$. The parameter $\gamma$ controls the strength of the modulating term in focal loss~\cite{lin2017focal} and is set to the same value as that in $L_{cls}$ directly. For $\alpha$, we perform a parameter sweep and use 0.5 for default setting. We refer readers to Section~\ref{as} for more details. Our models are trained on 4 V100 GPUs for 90K iterations with a minibatch of 16 images. Stochastic gradient descent (SGD) is adopted with the initial learning rate of 1e-2. The learning rate drops by $10{\times}$ at iteration 60K and 80K. Weight decay and momentum are 0.0001 and 0.9, respectively. The backbone network is initialized with weights pretrained on ImageNet\cite{deng2009imagenet} and other layers are initialized as in \cite{lin2017focal}.

\subsection{Ablation Study} \label{as}
In this section, we conduct ablation studies to evaluate our designed components, the coarse-to-fine module~(\emph{CFM}) and the holistic boundary-aware branch~(\emph{HBB}). Besides, more settings about our network are also discussed.

\begin{table}[htbp]
\caption{Results of Ablation Study}
\begin{center}
\begin{tabular}{c|c|c|c|c|c|c}
\hline
&\textbf{AP}&\textbf{AP$_{50}$}&\textbf{AP$_{75}$}&\textbf{AP$_{S}$}&\textbf{AP$_{M}$}&\textbf{AP$_{L}$} \\
\hline
\textbf{Baseline} & 30.4 & 51.1 & 31.2 & 13.5 & 33.5 & 43.9 \\
\textbf{+ CFM}  & 31.2 & 51.6 & 32.1 & 13.9 & 33.9 & 45.7 \\
\textbf{+ HBB} & 31.7 & 52.4 & 33.0 & 14.0 & 34.8 & 46.5\\
\hline
\end{tabular}
\label{ablation study}
\end{center}
\end{table}

\subsubsection{Coarse-to-Fine Module} 
The results of ablation studies are reported in Table~\ref{ablation study}. The first row~\emph{Baseline} reports the result of \cite{xie2020polarmask}, which is also the coarse module of our approach. The second row~\emph{CFM} presents the performance of our coarse-to-fine regression module. The coarse-to-fine architecture outperforms the baseline by 0.8 AP. Specifically, for different IOU thresholds, the AP$_{75}$ improvement 0.9 is greater than AP$_{50}$ improvement 0.5. This indicates that our coarse-to-fine module contributes to segmentation of details and is consistent with the mechanism of regressing in the local area accurately. Moreover, 1.8 AP$_L$ improvement verifies the coarse-to-fine module's powerful ability of long-distance regression. Fig.~\ref{corase_to_fine_results} shows qualitative results of the coarse-to-fine module and the baseline, where the coarse-to-fine module gives better boundaries.

\subsubsection{Holistic Boundary-Aware Branch} 
Comparing the second and the third row in Table~\ref{ablation study}, the holistic boundary-aware branch yields 0.5 AP improvement, which demonstrates the effectiveness of incorporating global information and image-level supervision. We can see the branch improves AP$_S$ slightly. It is because the boundary masks of small objects degrade more when downsampling. However, at least the holistic boundary-aware branch does not lower on AP$_S$.

\begin{table}[htbp]
\caption{Results of using different convolution in the refined regressor}
\begin{center}
\begin{tabular}{c|c|c|c|c|c|c}
\hline
&\textbf{AP}&\textbf{AP$_{50}$}&\textbf{AP$_{75}$}&\textbf{AP$_{S}$}&\textbf{AP$_{M}$}&\textbf{AP$_{L}$} \\
\hline
\textbf{\makecell{Grouped\\Conv 1$\times$1}}  & 31.2 & 51.6 & 32.1 & 13.9 & 33.9 & 45.7 \\
\textbf{\makecell{Standard\\Conv 1$\times$1}} & 31.3 & 51.8 & 32.4 & 13.7 & 34.4 & 45.5\\
\hline
\end{tabular}
\label{conv type}
\end{center}
\end{table}

\subsubsection{More Discussions}
As described before, the regressor in the fine module consists of a single convolutional layer. Here we study more about different convolution operations. With grouped Conv 1$\times$1, features are divided into 36 groups to convolve separately, which means there is no communication between points' features when regressing. As reported in Table~\ref{conv type}, standard convolution achieves slightly better performance than grouped convolution. The results show one boundary point benefits from other initial points' features quite limitedly in the fine regression stage, implying that the fine regression performs in a local area. Therefore, we use grouped convolution for all experiments to reduce computational complexity.

\begin{table}[htbp]
\caption{Comparison of different supervision forms on the initial module}
\begin{center}
\begin{tabular}{c|c|c|c|c|c|c}
\hline
&\textbf{AP}&\textbf{AP$_{50}$}&\textbf{AP$_{75}$}&\textbf{AP$_{S}$}&\textbf{AP$_{M}$}&\textbf{AP$_{L}$} \\
\hline
\textbf{\makecell{Implicit\\supervision}}  & 30.8 & 51.4 & 32.0 & 13.7 & 33.8 & 44.8 \\
\textbf{\makecell{Explicit\\supervision}} & 31.2 & 51.6 & 32.1 & 13.9 & 33.9 & 45.7\\
\hline
\end{tabular}
\label{initial loss}
\end{center}
\end{table}
Besides, we compare explicit and implicit supervision on the coarse regression stage. For explicit supervision, other than the fine regression loss, a coarse regression loss term is applied to directly penalize the deviation between the coarse prediction and ground truth. With implicit supervision, the coarse regression loss is dropped. Table~\ref{initial loss} shows that an explicit manner boosts the performance. We explain that the explicit form allows better initial prediction, especially at the early stage of training, which contributes to feature sampling in the fine stage.      

\begin{figure}[htbp]
\centerline{\includegraphics[scale=0.55,trim=80 270 100 280]{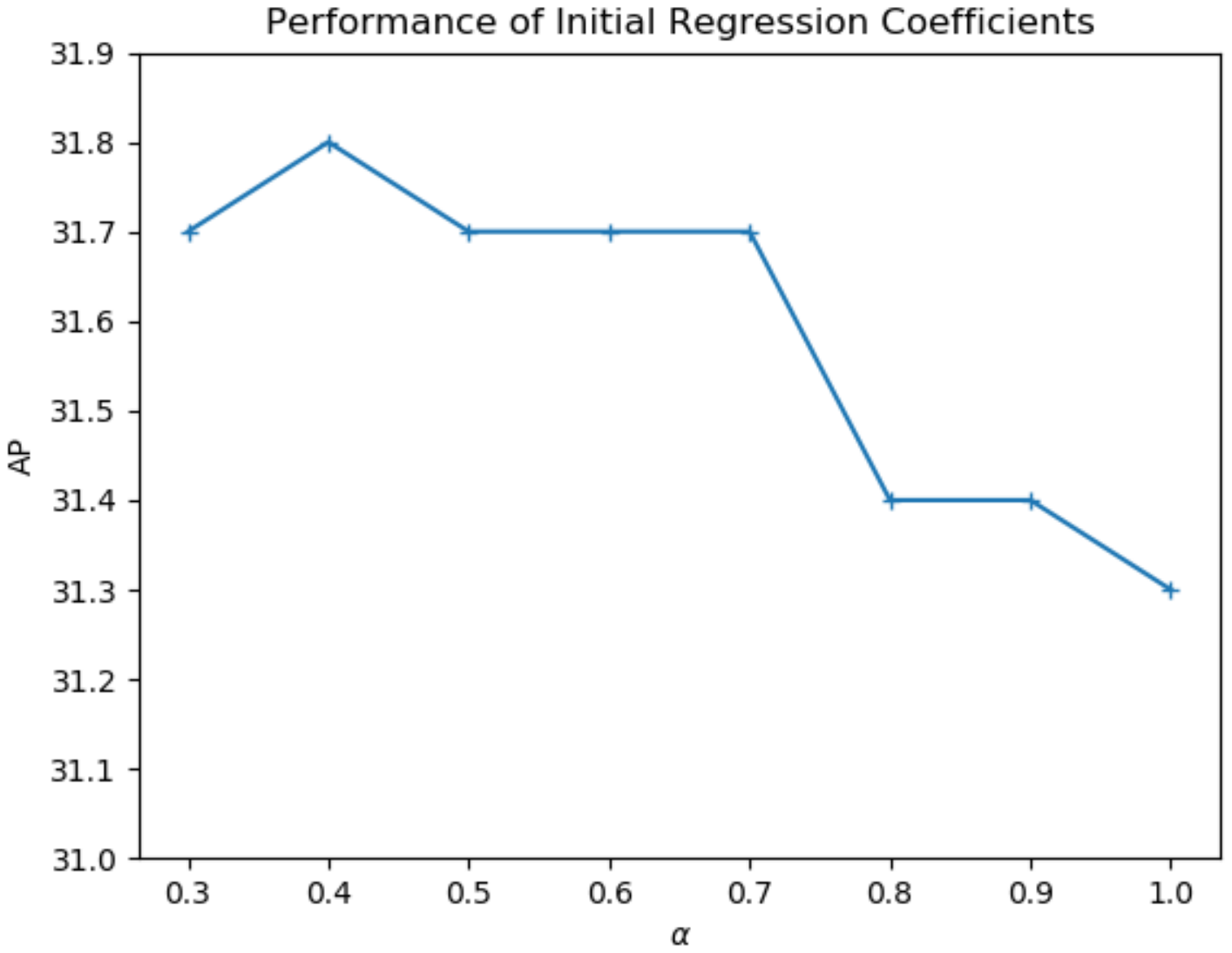}}
\caption{Hyper-parameter analysis on the ration of radius regression loss coefficients. The results show that the loss of fine regression module $L_{refine}$ should be weighted more than that of the coarse regression module $L_{coarse}$.}
\label{hyperparameter}
\end{figure}

Furthermore, to explore how the ratio of initial regression loss $L_{coarse}$ and refined regression loss $L_{fine}$ influences performance, we ablate the hyper-parameters in Eq.~\eqref{loss}. Let $\alpha$ range from 0.3 to 1 in steps of 0.1. As shown in Fig.~\ref{hyperparameter}, the performance degrades when the weight of $L_{coarse}$ is greater than 0.7. We think it is reasonable because output of the refined module is the final prediction and its loss deserves more weight than that of the coarse module. Note that the performance with $\alpha$ being 0.4 surpasses others in the range of [0.3, 0.7] very slightly, thus we ignore the minor superiority and use 0.5 for $\alpha$.

\begin{table}[htbp]
\caption{Computational Complexity and Parameters Analysis}
\begin{center}
\begin{tabular}{c|c|c|c|c}
\hline
\textbf{Method}&\textbf{Backbone}&\textbf{Size}&\textbf{GFLOPs}&\textbf{Params} \\
\hline
\textbf{PolarMask} & ResNet-101 & 800$\times$1200 & 328.78 & 53.40M\\
\textbf{Ours (w/o HBB)} & ResNet-101 & 800$\times$1200 & 330.27 & 53.41M\\
\textbf{Ours (w HBB)} & ResNet-101 & 800$\times$1200 & 337.37 & 53.86M\\
\hline
\end{tabular}
\label{computaion complexity}
\end{center}
\end{table}

\subsection{Computational Complexity and Parameters Analysis}
\begin{figure*}[ht]
\centerline{\includegraphics[scale=0.27]{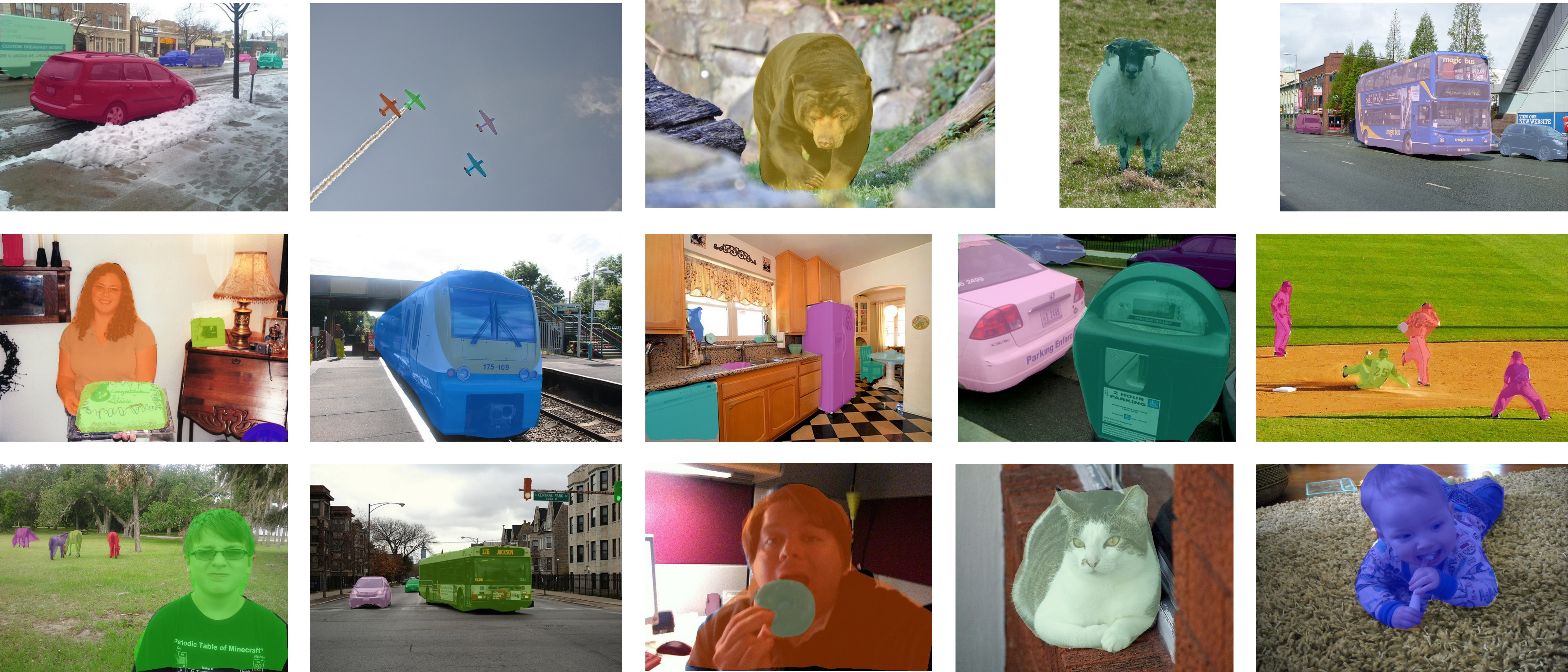}}
\caption{Quality results of our method on COCO \texttt{test-dev}, achieving
 31.7\% mask AP. Masks of multiple instances from the same image are shown in various colors for differentiation. (Best viewed in color.)}
\label{quality_results}
\end{figure*}

\begin{table*}[ht]
\caption{Quantitative Results on MS COCO \texttt{minival} and \texttt{test-dev}}
\begin{center}
\begin{tabular}{c||c|c|c||c|c|c|c|c|c|c}
\hline
\textbf{Method} & \textbf{Backbone} & \textbf{Epoch} & \textbf{Aug} & \textbf{AP$_{val}$} &\textbf{AP}&\textbf{AP$_{50}$}&\textbf{AP$_{75}$}&\textbf{AP$_{S}$}&\textbf{AP$_{M}$}&\textbf{AP$_{L}$} \\
\hline
ExtremeNet & Hourglass-104 & 100 & Yes & - & 18.9 & 44.5 & 13.7 & 10.4 & 20.4 & 28.3\\
ESE-Seg & Darknet-53 & 300 & No & 21.6 & - & - & - & - & - & -\\
DeepSnake & DLA-34 & 160 & Yes & 30.5 & 30.3 & - & - & - & - & -\\
PolarMask & ResNet-101-FPN & 12 & No & 30.4 & 30.4 & 51.9 & 31.0 & 13.4 & 32.4 & 42.8\\
\textbf{Our approach} & ResNet-101-FPN & 12 & No & \textbf{31.7} & \textbf{31.7} & 52.6 & 32.6 & 13.6 & 33.8 & 44.9\\  				
\hline
\end{tabular}
\label{quantitive results}
\end{center}
\end{table*}

We analyze the computational complexity and parameters of our designed modules. Results are reported in Table~\ref{computaion complexity}. In these experiments, ResNet-101 is used as backbone and the input image size is set as 800$\times$1200. As mentioned before, the holistic boundary-aware branch is dropped when testing, so we perform the analysis during training and testing separately. The second line in Table~\ref{computaion complexity} reports that our inference model only introduces 1.54 GFLOPs and 0.01M parameters additionally, which is negligible. Even if with the holistic boundary-aware branch, the number of parameters and GFLOPs increase marginally, demonstrating the lightweightness of our approach.

\subsection{Performance Comparison with State-of-the-Arts}
We compare the proposed approach with state-of-the-art boundary-based methods on COCO \texttt{test-dev}, reported in Table~\ref{quantitive results}. The performance of these previous works is available from published papers. Since some methods only provide performance on COCO 2017 val, our validation results are also reported. Here~\emph{aug} means data augmentation, including random crop and multi-scale training. Without bells and whistles, our approach can achieve competitive performance with all other boundary-based methods. It surpasses PolarMask~\cite{xie2020polarmask} on \texttt{test-dev} again with mask AP 31.7\%, which shows the robustness and reliability of our method. Moreover, with simple end-to-end training, our method outperforms Deep Snake~\cite{peng2020deep}. The qualitative results are visualized in Fig.~\ref{quality_results}.

\section{Conclusion}
In this work, we propose a coarse-to-fine regression module and a holistic boundary-aware branch for instance segmentation. The coarse-to-fine regression module decomposes difficult long-distance regression into two simple steps: approximate prediction in the global image and refined prediction in the local area. Furthermore, the holistic boundary-aware branch with image-level supervision guides the network to pay more attention to boundaries in the whole image context and assists radius regression explicitly. With single scale training and testing, our model achieves 31.7\% mask AP on COCO~\texttt{test-dev}. Due to its simplicity and effectiveness, the proposed method is quite competitive among existing boundary-based instance segmentation approaches.

\bibliographystyle{IEEEtran}
\bibliography{reference}
\end{document}